# A Data Fusion Platform for Supporting Bridge Deck Condition Monitoring by Merging Aerial and Ground Inspection Imagery

Zhexiong Shang,[1] Chongsheng Cheng,[2] and
Zhigang Shen, Ph.D.[3]

[1]NH 113, Durham School of Architectural Engineering & Construction, University of Nebraska-Lincoln, Lincoln, NE, 68588; e-mail: szx0112@huskers.unl.edu
[2]NH 113, Durham School of Architectural Engineering & Construction, University of Nebraska-Lincoln, Lincoln, NE, 68588; e-mail: cheng.chongsheng@huskers.unl.edu
[3]NH 113, Durham School of Architectural Engineering & Construction, University of Nebraska-Lincoln, Lincoln, NE, 68588; e-mail: shen@unl.edu

## ABSTRACT

UAVs showed great efficiency on scanning bridge decks' surface by taking a single shot or through stitching a couple of overlaid still images. If potential surface deficits are identified through aerial images, subsequent ground inspections can be scheduled. This two-phase inspection procedure showed great potentials on increasing field inspection productivity. Since aerial and ground inspection images are taken at different scales, a tool to properly fuse these multi-scale images is needed for improving the current bridge deck condition monitoring practice. In response to this need a data fusion platform is introduced in this study. Using this proposed platform multi-scale images taken by different inspection devices can be fused through geo-referencing. As part of the platform, a web-based user interface is developed to organize and visualize those images with inspection notes under users' queries. For illustration purpose, a case study involving multi-scale optical and infrared images from UAV and ground inspector, and its implementation using the proposed platform is presented.

## INTRODUCTION

Bridge deck serves as the key load-carrying structure throughout the service life of bridges. It constantly exposed to traffic, scour and chemicals that cause surface cracks and invisible corrosions of reinforcing steels in bridge decks. Traditional non-destructive evaluation (NDE) methods such as visual surveying, chain-dragging and hammer sounding, penetrating radar (GPR), impact echo (IE), ultrasound are labor intensive and time consuming. Following the significant advancement in computer vision and image processing, direct hyper-spectrum scanning (e.g. optical and IR camera) has demonstrated great potentials for detecting both surface and subsurface defects (Oh et al. 2012). Compared to other NDE methods, the major advantages of hyper-spectrum scanning are its high-speed data collection and its ease of locating and mapping the detected defects (Graybeal et al. 2002).

     Following the boom of UAVs, there has been significant growths of aerial inspection on bridge decks (Ellenberg et al. 2016; Metni and Hamel 2007). UAV has multiple advantages over the traditional ground-level inspection units. The low-cost UAVs can rapidly scan bridge decks with no need of traffic closure, and access regions that are dangerous and hard to be reached by

human. Equipped with onboard camera, UAVs can provide orthogonal views of bridge deck surface that would be otherwise inaccessible. In practice, UAVs need to fly at certain safe height above the deck surface to avoid the traffic and site obstacles during routine inspections (Zink and Lovelace 2015).

Lately, engineers in NDE community start to deploy a UAV-based two-phase inspection approach (Khan et al. 2015), with Phase 1 including the driving speed of UAV to cover the entire deck surface, followed by higher resolution inspection methods carried out on a subset of the decks based on the Phase 1 results and agency needs. This two-phase bridge inspection strategy showed great potentials on increasing field inspection productivity and, at the same time, reduce unnecessary cost. Due to the fact that aerial and ground images are taken at different scales, fusing these images through pure image-based technique is challenging (Fruh and Zakhor 2003).

An alternative method is to use the data management tool to support this multi-scale image data fusion. In the study, a multi-scale image data fusion platform is developed to support this newly emerged two-phase bridge deck inspection practice. The platform integrated the image acquisition process with the post-processing step for data preparation. A web-based user interface is then applied to query and visualize the captured images based on the user interaction. A real project is utilized to demonstrate the practicability of this platform.

**METHODOLOGY**

Fig. 1 shows the overview of the proposed platform, it includes the in field data collection with both aerial and ground inspection units, image processing for bridge deck surface mapping and the web-based user interface (UI). The web-based UI is served as a data management tool that properly organize the aerial and ground inspection to support both on-site bridge deck inspection and off-site condition assessment practice. Details of each section of the platform are discussed in the followings.

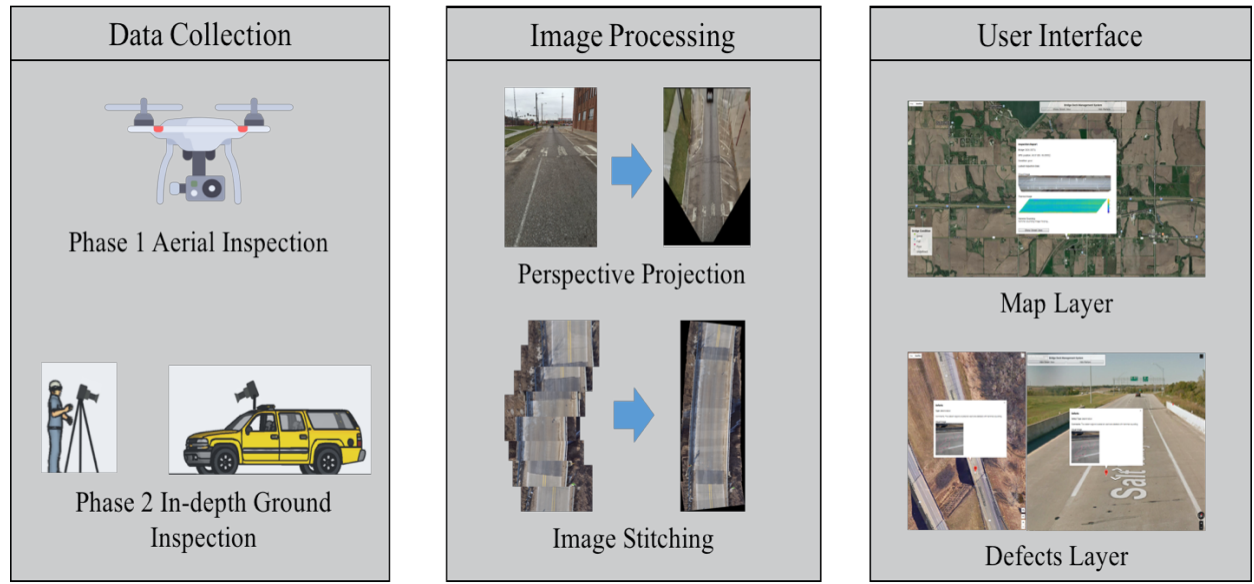

**Figure 1 Overview of the data collection, image processing and the user interface of the platform**

**Data Collection**
The data collection process is determined based on the inspection phases. For phase 1 inspection, images are captured by UAV scans along the bridge deck surface with fixed velocity and height. It is noted that the height is determined by the required spatial resolution and field of view (FOV) of onboard camera. Compared to the optical camera, IR camera has narrow FOV where larger distance is required to cover the entire deck width with single shot. During aerial inspection, the onboard camera is oriented towards the ground and the images are taken at short time internal so that sufficient overlay for mosaic map generation is guaranteed. The images captured in the air are geo-tagged and stored in a MicroSD card for post processing. Figure 3 shows the example of using UAVs (i.e. DJI Inspire 1) to scan bridge deck surface. For onboard camera has small FOV or UAV cannot reach the level of height because of field obstacle (e.g. truss, beams), assigning a gimbal pitch angle to the camera can still reach the full deck coverage. However, additional post-processing is required to eliminate the camera's perspective effect.

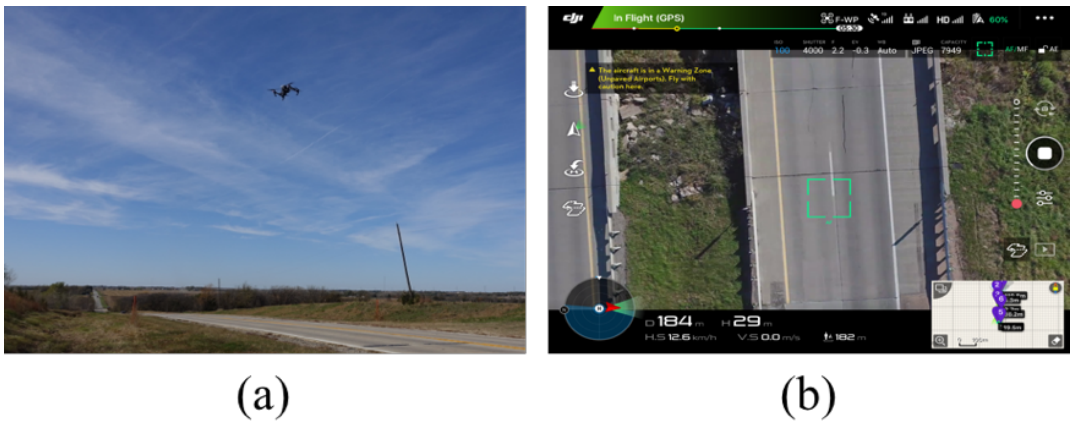

**Figure 2 (a) Field view of aerial inspection process; (b) Camera view of image acquisition progress**

If potential defects are identified in the aerial-image generated map, the subsequent in-depth inspection is then carried out by ground inspection units. In general, the subsequent inspection is only applied at the potential defects that are detected in aerial inspection. However, for bridge deck that is heavily deteriorated, full deck inspection with inspectors or ground vehicles is high demanded.

**Image Processing**
Images collected during inspection are either single image which indicate specific defects verified in the subsequent inspection, or a set of overlaid images for bridge deck surface mapping. In this study, two image processing techniques are utilized to support the construction of orthogonal surface map of bridge deck, they are: perspective projection and image stitching. Perspective projection is the technique to remove the perspective effects in inspection images. For bridge deck surface mapping, images not taken with orthogonal view can result in misinterpretation of the real dimensions and shapes of the defects in the constructed map. Such issue is often existed for aerial and ground inspection process where the onboard camera's shooting angle is not perpendicular to the deck surface.

In this study, inverse perspective mapping (IPM) is used to remove such effect by re-projecting the distorted images into an orthogonal view. IPM is a geometrical transformation algorithm that converts an incoming image at a 2D space into the 3D world coordinate, and remaps each pixel toward a different position in 3D and construct a new 2D image plane with corrected view (Mallot et al. 1991). The transformation result is a new image with bird eye view of the original image at significant height (as shown in Fig. 3). The pixel-wise formula of IPM is presented in Equation (1).

$$u = \frac{\text{atan}\left(\frac{h sin \gamma(x,y)}{y-d}\right) - (\theta - \alpha)}{\frac{2\alpha}{m-1}}$$
$$v = \frac{\text{atan}\left(\frac{y-d}{x-l}\right) - (\gamma - \alpha)}{\frac{2\alpha}{n-1}}$$
(1)

Where (l, d, h) is camera position in the world coordinates, (θ, γ) is the pitch and yaw angles of camera. α is camera aperture, and (m, n) is the image resolution. (x, y) is the pixel coordinates in the original image, and (u, v) is the pixel coordinates in the corrected image. For bridge deck inspection, the camera position can be set above the origin of world coordinate (as $l, d = 0$) with camera yaw aligns with the front axis ($\gamma = 0$), which simplifies its calculation process. Fig. 4 (a) and (b) show the example of a thermal image with tilted pitch angle before and after re-projection using IPM.

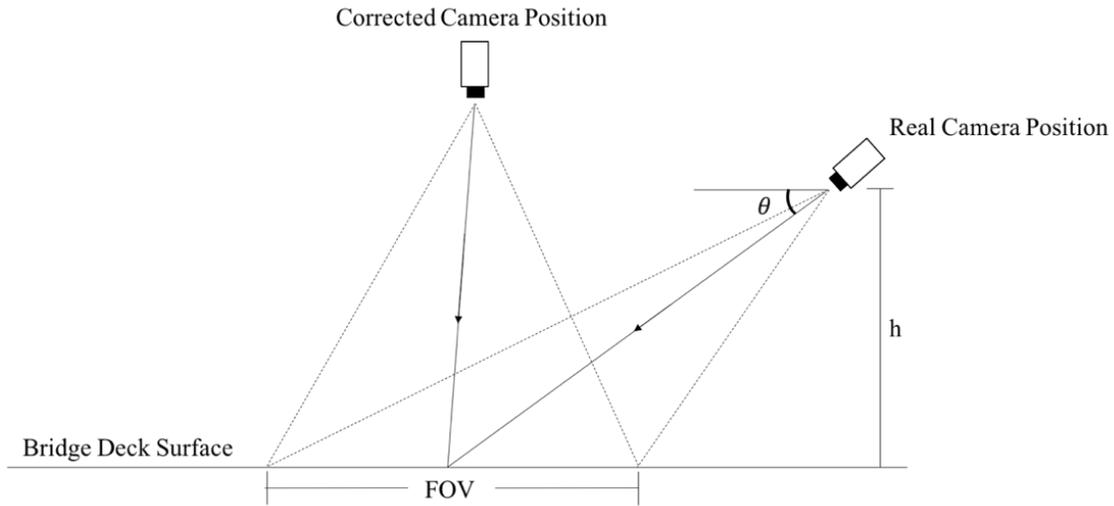

**Figure 3 Conceptual diagram of IPM**

The next step is to stitch these corrected images in order to construct bridge deck surface map. There is a wide variety of algorithmic solutions available for stitching the overlaid images. SIFT/SURF feature detectors are widely used to identify salient features within each images, and matching those features in adjacent images (Szeliski 2007). Image transformation is then applied to align the matched images into a uniform coordinate system. This strategy works well for most image sets where salient features can be detected and matched. However, especially for thermal images where salient features may not always exist, directly applying the feature-based matching may arise significant errors. In this platform, image stitching is a semi-autonomous process where the feature based automatic image matching is applied as the initial step (Brown and Lowe 2007). For images without enough salient features, the manual process is taken over where the GPS locations attached on each image are used as the reference for image registration.

Fig. 4 (c) shows the surface map constructed with the set of corrected IR images using the semi-autonomous image stitching strategy. The real dimension and shapes of defects can then be revealed in the constructed map.

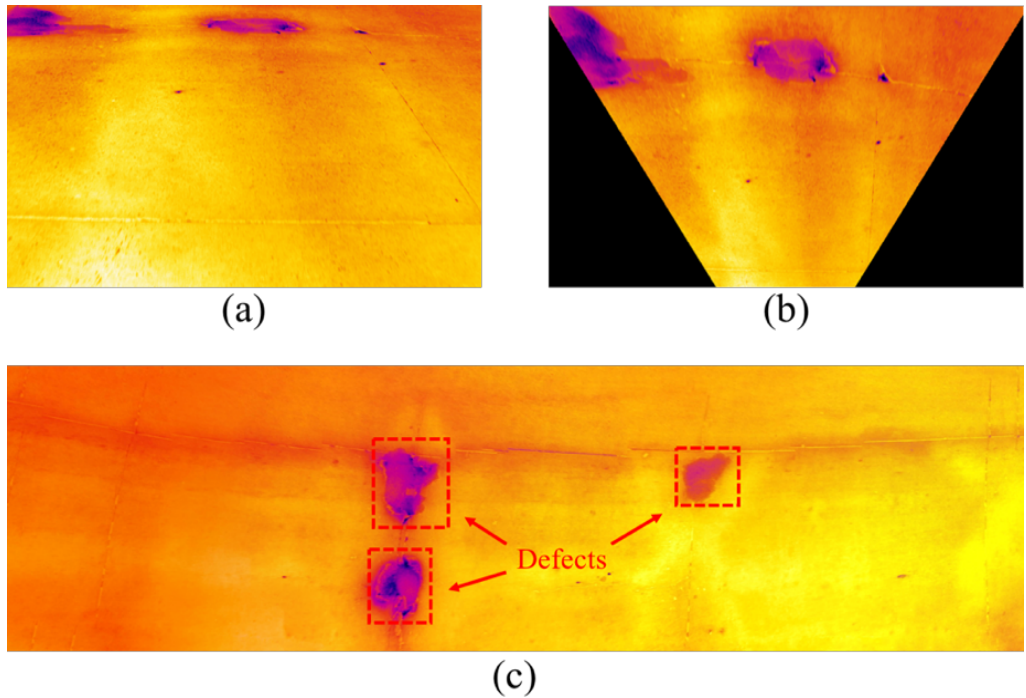

**Figure 4 (a) IR image with perspective effect; (b) Image corrected with IPM; (c) Stitching the corrected images to identify real dimension of defects.**

**User Interface**
In this section, a web-based user interface is introduced to properly merge and manage the images taken with different inspection phases. The user interface is implemented with JavaScript and Google Map API which synchronize images at different scales into the uniform GPS coordinate system. To support the user interaction and data management, two data layers are developed in this UI, they are: map layer and defects layer. The detailed discussion of each data layer is presented below.

Map layer is served as the base holder of this UI. It provides the outline map view which allows user to access bridge decks by mouse selection in Google Map. The aerial inspection constructed surface maps are linked to this data layer through the GPS coordinates. Due to the geometrical transformation of the mosaic image set, GPS of the initial image of the map is used to localize such surface maps. Callback functions are also developed in this layer to visualize the condition of individual bridge deck as well as the stored surface maps through a pop-up window. Dataset stored in the map layer are mostly applied by bridge owners who are more interested in the the overall conditions of bridge decks within the region of interest (ROI) and the general inspection results of each bridge deck.

Defects layer is the place holder to store the inspection notes and images that indicates specific bridge defects detected in the ground-level inspection. These defects are normally captured in the surface map and further validated in the subsequent inspections. Images in this layer can be either image data that directly capture the surface cracks (e.g. by optical camera) and subsurface delamination (e.g. by IR camera), or the images of markers that represents the defects detected by other sensors (e.g. GPR, ultrasound, hammer sounding). For individual images, the geotag attached on each image is used to locate the defects position in the map.

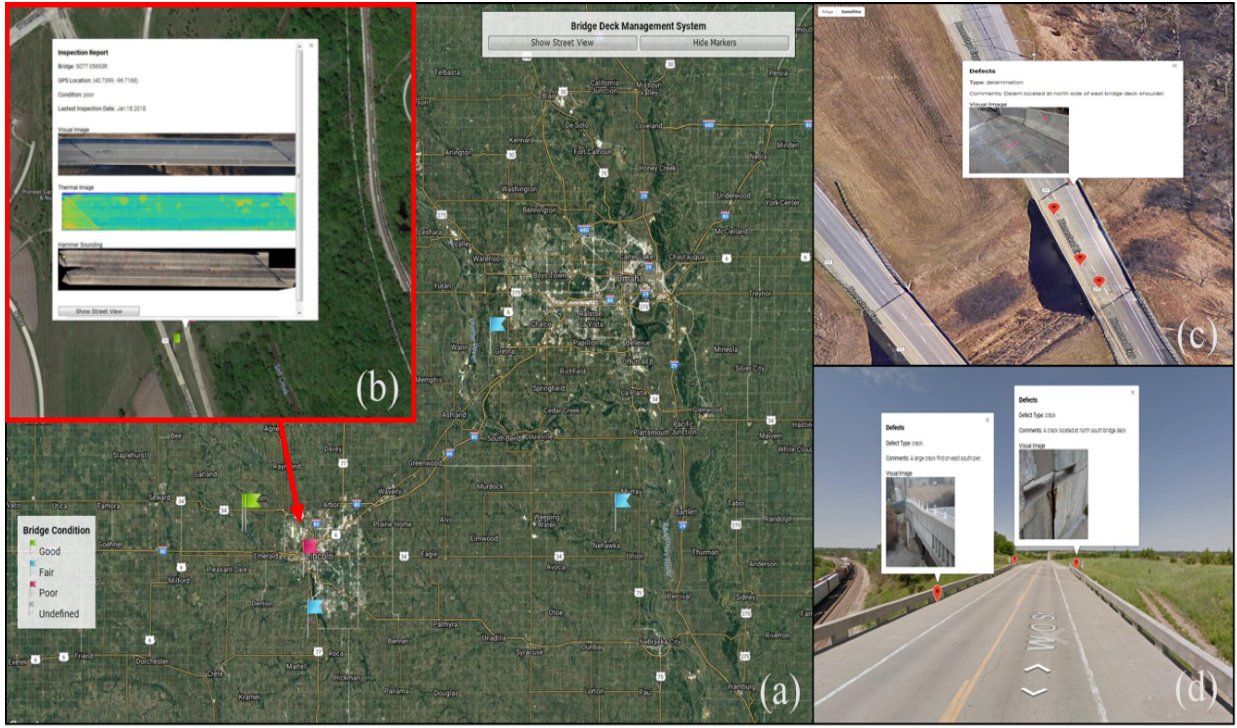

**Figure 5 Web-based User Interface: (a) Map view to indicate general conditions of bridges within the ROI; (b) Click each flag shows the pop-up window of the conditions as well as the surface map of the selected bridge deck; (c) Zoom in the map to show the locations of bridge deck defects (i.e. delamination) with captured image and description in subsequent inspection; (d) Street view of bridge deck defects (i.e. surface crack) for assisting in identifying defects locations on-site.**

Five bridges located near Lincoln, Nebraska are surveyed by Nebraska Department of Transportation (NDOT), inspection images, including both aerial images captured by UAV and ground hammer sounding with handhold camera, as wells as the on-site notes were uploaded into the platform after the field inspection period. In this section, the bridge deck with significant structural deficient (shown as red flag in Fig. 5) is chosen to demonstrate the functionality and the practical usage of the UI. In Fig. 5 (a), the general conditions of bridge decks within the region of interest (ROI) are presented in a 2D map view where the condition of each bridge is illustrated with a colored flag. Clicking each flag in the map shows the pop-up window where the general condition and constructed surface map of the bridge deck are presented. Fig. 5 (b) shows an example of the use case where the displayed surface maps captured by UAV are presented. In this example, three surface maps are presented, they are: surface map constructed by phase 1 aerial inspection using digital camera (top), surface map constructed by phase 1 aerial inspection using IR camera (middle), surface map constructed by phase 2 ground inspection using hammer sounding. By zooming in the map view to meter level, individual defects identified through handhold camera and hammer sounding through phase 2 inspection are automatically displayed. Clicking each marker in the zoomed view displays the image and description of identified defects. Fig 5 (c) shows the example where the locations, comments as well as the defects focused inspection image are presented. In order to support the on-site defects identification, the platform also provides a function which allows users to localize the defects in the street view (as shown in Fig 5 (d)). This street view can support engineers staying on-site to identify the locations of the subsurface defects (e.g. delamination) without the need of manually drawing the reference coordinates on the deck surface. Noted that the accuracies of street view images are highly determined by the precision of GPS signal and the updates of the google map. Advanced image processing and sensor fusion algorithms can increase such accuracy, however, discussing such technologies is out of the scope of this study.

## SUMMARY

In this study, a web-based data fusion platform is introduced to support merging multi-scale bridge deck inspection images through geo-referencing. The platform is compatible with the newly emerged UAV-enabled, two-phase bridge deck inspection practice. The data collection, image processing and visualization process of using the proposed platform on such practice are discussed. With the increased flight capability and the resolution of onboard camera system, the role of UAV on structural inspection practice is emerging. In the future studies, the authors aim to expand the proposed platform to support data management of versatile UAV-enabled inspection practice, such pavements or tunnels.

## KNOWLEDGEMENT

The image acquisition process received support from the Nebraska Department of Transportation through facilitating data collection and sharing the ground evaluation results.